\documentclass{article}

\usepackage{arxiv}
\usepackage[utf8]{inputenc} 
\usepackage[T1]{fontenc}    
\usepackage{hyperref}       
\usepackage{url}            
\usepackage{booktabs}       
\usepackage{amsfonts}       
\usepackage{nicefrac}       
\usepackage{microtype}      
\usepackage{lipsum}
\usepackage{graphicx}

\usepackage{amsmath}
\DeclareMathOperator*{\argmin}{arg\,min}
\usepackage{amssymb}
\usepackage{stmaryrd}
\usepackage{cmap}
\usepackage{subcaption}

\usepackage{algorithm}
\usepackage{algpseudocode}
\algnewcommand\algorithmicinput{\textbf{Input:}}
\algnewcommand\INPUT{\item[\algorithmicinput]}
\algnewcommand\algorithmicoutput{\textbf{Output:}}
\algnewcommand\OUTPUT{\item[\algorithmicoutput]}
\graphicspath{ {./images/} }

\title{CNN Acceleration by Low-rank Approximation with Quantized Factors}

\author{
 Nikolay Kozyrskiy \\
  Skolkovo Institute of Science and Technology\\
  Moscow, Russia \\
  \texttt{nikolay.kozyrskiy@skoltech.ru} \\
   \And
 Anh-Huy Phan \\
  Skolkovo Institute of Science and Technology\\
  Moscow, Russia \\
  \texttt{a.phan@skoltech.ru} 
}

\begin{document}
\maketitle
\begin{abstract}
The modern convolutional neural networks although achieve great results in solving complex computer vision tasks still cannot be effectively used in mobile and embedded devices due to the strict requirements for computational complexity, memory and power consumption. The CNNs have to be compressed and accelerated before deployment. In order to solve this problem the novel approach combining two known methods, low-rank tensor approximation in Tucker format and quantization of weights and feature maps (activations), is proposed. The greedy one-step and multi-step algorithms for the task of multilinear rank selection are proposed. The approach for quality restoration after applying Tucker decomposition and quantization is developed. The efficiency of our method is demonstrated for ResNet18 and ResNet34 on CIFAR-10, CIFAR-100 and Imagenet classification tasks. As a result of comparative analysis performed for other methods for compression and acceleration our approach showed its promising features.
\end{abstract}


\section{Introduction}
\hspace{10mm} The modern deep convolutional neural networks (CNNs) \cite{LeCun1989BackpropagationAT, he2015deep, simonyan2014deep, ronneberger2015unet} have proved themselves to be powerful tools for solving various complicated tasks, such as image classification \cite{touvron2020fixing}, semantic segmentation \cite{yuan2019objectcontextual, chen2018encoderdecoder}, object detection \cite{zhang2020resnest}, image enhancement \cite{zhang2018image, zhou2019awgnbased}, image generation \cite{karras2018stylebased} and many others. 

\hspace{10mm} A number of flexible high-level deep learning frameworks for conducting experiments with neural networks have been created so far \cite{abadi2016tensorflow, paszke2019pytorch}. The powerful but rather big, expensive and energy-demanding graphics processing units (GPUs), which are efficient for fast matrix and tensor computations due to their architecture, have also become widely available. With the help of the mentioned tools researchers have developed well-performing but large and expensive to execute models \cite{Khan_2020}. Deployed in the cloud with ample computing power, such deep cumbersome neural networks have become a substantial part of many real-world applications including automatic translation systems, voice assistants, advanced recommender systems, search engines and image processing tools, just to name a few. 

\hspace{10mm} The advantages of modern CNNs can also be very useful for mobile applications. However the mobile devices have strong restrictions on computation complexity, power and memory consumption. It is essential to compress and speed up the CNNs prior to deployment on a mobile device in order to meet the hardware requirements. Here the compression term refers to reducing the amount of memory required to store the CNN. The speed up term in its turn refers to decreasing the inference time. 

\hspace{10mm} In fact not only mobile devices are in dire need of CNNs compression and speeding-up but all on-line applications where the low latency is vital. Autopilots for autonomous driving \cite{Grigorescu_2020}, video stream processing \cite{Guan_2019} are among such applications, for example.

\hspace{10mm} Deep convolutional neural networks are typically over-parameterized, which contributes to the convergence to a good local minima of the loss function at the training stage \cite{allenzhu2018learning, denil2013predicting}. However such over-parametrization makes CNNs too large and slow to compute so that in order to speed up the inference stage and decrease the memory consumption the redundancy can be eliminated. Generally the compression and acceleration methods make sense in case of insignificant quality drop, still it depends on the task and the requirements.

\hspace{10mm} The approaches for compression and acceleration of CNNs can be summarized into the following groups: parameters pruning\cite{Cun90optimalbrain, han2015learning}, weights sharing\cite{chen2015compressing}, quantization of weights and activations \cite{gong2014compressing, wu2015quantized}, low-rank tensor and matrix factorization \cite{lebedev2014speedingup, kim2015compression}, knowledge distillation \cite{hinton2015distilling}. The idea behind parameters pruning and weights sharing in general is based on removing redundant parameters which are insensitive to the CNN performance. Quantization here is the synonym of discretization. The essence of this approach is reducing the number of bits required to represent the parameter or activation in order to take advantages of fixed-point arithmetic. The low-rank approximation approach allows to remove the redundancy of order-4 tensors via a low-rank tensor decomposition for convolutional layers and of matrices via a low-rank matrix decomposition for fully-connected layers. The main idea of knowledge distillation methods is to compress the powerful but cumbersome teacher model to the shallower one (student), which would be able to reproduce the function learned by the teacher. The application of the particular method is determined by the given task and CNN structure. Some of the mentioned approaches can often be applied simultaneously in order to achieve best compression and acceleration ratios \cite{han2015deep}.

\hspace{10mm} In the current work the approach of combining low-rank factorization and quantization methods is considered. As the quantization uses the redundancy in parameters representation while the low-rank approximation uses the redundancy of over-parametrization, these two parts are in some sense orthogonal.
This is why there are reasons to expect that the joint application of these two approaches can produce tangible compression without noticeable quality drop. Moreover the approach of combining tensor decomposition with quantization for acceleration is quite novel by now and there is no exhaustive research in this field yet.

\hspace{10mm} The most famous tensor decompositions are Canonical Polyadic (CPD) \cite{sapm19287139}, Tucker decomposition \cite{Tuck1966c} and tensor-train decomposition \cite{OseledetsTT}. The Tucker tensor format is chosen for the low-rank approximation of convolutional weights. The quantization is applied to both weights and activations. The experiments with four, five, six, seven and eight bits for fixed-point representation are conducted. The cases of extreme quantization - binary and ternary - are not considered in this work as they require different approach \cite{rastegari2016xnornet}. The algorithms and methods for ranks selection and quality restoration are proposed. ResNet18 and ResNet34 for classification tasks on CIFAR-10, CIFAR-100 and Imagenet datasets are chosen for experiments. As a result we present the method for compression and acceleration with the ability to control the tradeoff between compression ratio and quality drop.

\section{Related Work}
\hspace{10mm} Convolutional layers make the biggest contribution to the most computations of deep CNNs. Thus optimizing the number of parameters and floating point operations (FLOPS) in convolutional filters results in good compression of the whole CNN.

\subsection{Low-rank Factorization}
\hspace{10mm} One of the first works on acceleration of CNNs with the help of low-rank approximations is \cite{denton2014exploiting}. The authors exploit the linear structure present within the convolutional filters in order to develop approximations which reduce the computations. The report the 2x speedup for a layer with 1\% accuracy drop on classification task. In \cite{jaderberg2014speeding} the cross-channel or filter redundancy was exploited for construction the low-rank basis of filters. The developed methods showed the 4.5x compression with less than 1\% drop in classification accuracy. Continuing to work in the similar direction, Lebedev et al. in their research \cite{lebedev2014speedingup} applied low-rank CPD on order-4 convolution kernel tensor. The experiments conducted in this paper demonstrated 8.5x CPU speedup with 1\% accuracy drop for the 36-class character classification CNN and 4x speedup of second layer of the AlexNet\cite{alexnet} for ImageNet classification \cite{russakovsky2014imagenet}. The paper \cite{kim2015compression} presents the one-shot whole CNN compression scheme based on the Tucker decomposition. Authors also describe the rank selection method with the help of variational Bayesian matrix factorization (VBMF) \cite{vbmf}. Another tensor decomposition - tensor-train decomposition - is used in \cite{alex2015tensorizing} for converting the weight matrices of the fully-connected layers to the TT-format. TT-decomposition is also used in \cite{garipov2016ultimate}. In this work both convolutional and fully-connected layers are compressed with TT-decomposition. The significant compression ratio of 80x is achieved with 1.1\% accuracy drop on the CIFAR-10 classification.

\subsection{Quantization}
\hspace{10mm} Quantization of weight and activations of convolutional and fully-connected layers is an efficient way of compressing and accelerating. The $k$-means scalar quantization is applied in \cite{gong2014compressing} and \cite{wu2015quantized}. In \cite{choi2016limit} the method based on the Hessian-weighted $k$-means for clustering network parameters is proposed. Authors of \cite{wu2015quantized} developed the framework for CNN quantization and described the non-uniform quantization scheme based on the dictionary learning. Anwar et al. \cite{anwarq} proposed the efficient method of training quantized CNNs for object recognition tasks. In the work \cite{jacob2017quantization} the scheme for integer-only arithmetic is provided. Jacob et al. use uniform quantization for efficient implementation in real hardware. They reported the comparative results in terms of latency-vs-accuracy tradeoff for floating-point and fixed-point arithmetic for MobileNets on ImageNet using the Qualcomm Snapdragon 835 cores. 

\hspace{10mm} The biggest speedup and compression theoretically can be achieved with binarizing of CNNs. However such extreme case of quantization requires more complex approaches compared to higher bitness. In \cite{courbariaux2016binarized} authors developed binarized CNN and replaced most arithmetic operations with bit-wise operations. The significant improvement in power-efficiency and inference time are reported. Rastegari et al. in \cite{rastegari2016xnornet} propose XNOR-Net, the CNN with binary weights and activations, and describe the training process. XNOR-Net outperformed BinaryNet by large margins on ImageNet dataset. In the recent work on binary CNNs \cite{bulat2019matrix} the training scheme based on SVD and Tucker decompositions is described. The linear or multi-linear over-parametrization of the CNN weights allowed to exploit the inter-dependency between the binary filters. The results provided in this work are the state-of-the-art for human pose estimation on MPII and image classification on ImageNet.

\section{Tucker Decomposition}
\subsection{Preliminaries}
\hspace{10mm} A tensor can be considered as a multiway numerical array. The order of a tensor indicates the number of its modes (dimensions). An order-$N$ tensor with real values is denoted by $\mathcal{A} \in \mathbb{R}^{I_1 \times I_2 \times \cdots \times I_N}$ with its entries $a_{i_1 i_2 \cdots i_N} \in \mathbb{R}$. A mode-$N$ product of a tensor $\mathcal{A} \in \mathbb{R}^{I_1 \times I_2 \times \cdots \times I_N}$ and a matrix $\mathbf{B} \in \mathbb{R}^{L_k \times I_k}$ is denoted $\mathcal{A} \times_N \mathbf{B}$ and results in a tensor $\mathcal{C} \in \mathbb{R}^{I_1 \times \cdots I_{k-1} \times L_k \times I_{k+1} \times \cdots \times I_N}$ with entries $c_{i_1 \cdots i_{k-1} l_k i_{k+1} \cdots i_N} = \sum\limits_{i_k=1}^{I_k} a_{i_1 \cdots i_{k-1} i_k i_{k+1} \cdots i_N} \cdot b_{l_k i_k}$. The mode-$N$ product is used to define the full multilinear product: $\mathcal{C} = \mathcal{A} \times_1 \mathbf{B}^{(1)} \times_2 \mathbf{B}^{(2)} \cdots \times_N \mathbf{B}^{(N)} = \llbracket \mathcal{A}; \mathbf{B}^{(1)}, \mathbf{B}^{(2)}, \cdots, \mathbf{B}^{(N)}  \rrbracket$.

\hspace{10mm} The Tucker decomposition (TKD) of an order-$N$ tensor $\mathcal{T} \in \mathbb{R}^{I_1 \times I_2 \times \cdots \times I_N}$ is denoted as follows: 
\begin{equation}
    \mathcal{T} = \llbracket \mathcal{G}; \mathbf{B}^{(1)}, \mathbf{B}^{(2)}, \cdots, \mathbf{B}^{(N)}  \rrbracket,
\end{equation}
where $\mathcal{G} \in \mathbb{R}^{R_1 \times R_2 \times \cdots \times R_N}$ is a core tensor, $\mathbf{B}^{(n)} \in \mathbb{R}^{I_n \times R_n} \Big|_{n=1}^{N}$ are factor matrices and the ordered set $(R_1, \cdots, R_N)$ is called the multilinear rank of the tensor $\mathcal{T}$. The multilinear rank is a natural extension of the matrix rank in the sense that $R_i$ is the dimension of the subspace spanned by mode-$i$ fibers. Unlike row and column ranks in matrix case (i.e. for $N=2$ $R_1 = R_2$) for $N \geq 3$ the values $R_1, \cdots, R_N$ can be different. The problem of determining the full multilinear rank is NP-hard in general.

\hspace{10mm} Tucker decomposition is not unique in general if unconstrained. At the same time the subspaces determined by the factor matrices are unique. 

\hspace{10mm} One of the common constraints for TKD is orthogonality \cite{spm}. In the work \cite{Lathauwer2000OnTB} the algorithm for best rank-$(R_1, R_2, \cdots,  R_N)$ approximation $\mathcal{\widetilde{T}}$ of tensor $\mathcal{T} \in \mathbb{R}^{I_1 \times I_2 \times \cdots \times I_N}$ with orthogonality constraints is derived. It is called higher-order orthogonal iteration (HOOI). The basic idea is the following. Given the tensor $\mathcal{T} \in \mathbb{R}^{I_1 \times I_2 \times \cdots \times I_N}$ find the low-rank approximation 
\begin{equation}
    \mathcal{\widetilde{T}} = \mathcal{B} \times_1 \mathbf{U}^{(1)} \times_2 \mathbf{U}^{(2)} \cdots \times_N \mathbf{U}^{(N)},
\end{equation}
which minimizes the objective function 
\begin{equation}
    f(\mathcal{\widetilde{T}}) = ||\mathcal{T} - \mathcal{\widetilde{T}} ||^2_F.
\end{equation}
Each factor matrix $\mathbf{U}^{(i)} \in \mathbb{R}^{I_i \times R_i} \big|_{i=1}^{N}$ has orthonormal columns and $\mathcal{B} \in \mathbb{R}^{R_1 \times R_2 \times \cdots \times R_N}$ is the core tensor. It is shown that for determined matrices $\mathbf{U}^{(1)}, \cdots, \mathbf{U}^{(N)}$ the core that minimizes the objective function $f(\mathcal{\widetilde{T}})$ is given by 
\begin{equation}
    \mathcal{B} = \mathcal{T} \times_1 \mathbf{U}^{(1)^T} \times_2 \mathbf{U}^{(2)^T} \cdots \times_N \mathbf{U}^{(N)^T}.
\end{equation}
The task of minimization $f(\mathcal{\widetilde{T}})$ is equivalent to maximization of 
\begin{equation}
    g(\mathbf{U}^{(1)}, \cdots,  \mathbf{U}^{(N)}) = ||\mathcal{T} \times_1 \mathbf{U}^{(1)^T} \times_2 \mathbf{U}^{(2)^T} \cdots \times_N \mathbf{U}^{(N)^T} ||^2_F .
\end{equation}
For fixed matrices $\mathbf{U}^{(1)}, \cdots, \mathbf{U}^{(n-1)},  \mathbf{U}^{(n+1)}, \cdots, \mathbf{U}^{(N)}$ the function $g$ is:
\begin{equation}
    g = ||\mathcal{\widetilde{U}}^{(n)} \times_n \mathbf{U}^{(N)^T}||^2_F,\; \text{where}
\end{equation}
\begin{equation}
    \mathcal{\widetilde{U}}^{(n)} = \mathcal{T} \times_1 \mathbf{U}^{(1)^T} \cdots  \times_{n-1} \mathbf{U}^{(n-1)^T}  \times_{n+1} \mathbf{U}^{(n+1)^T} \cdots \times_N \mathbf{U}^{(N)^T}.
\end{equation}
The columns of $\mathbf{U}^{(n)}$ are found as the orthonormal basis for the dominant subspace of the $n$-mode space of  $\text{ }\mathcal{\widetilde{U}}^{(n)}$. The same procedure can be applied to any modes of $ \mathcal{T}$ for the minimization of $f(\mathcal{\widetilde{T}})$.

The compression ratio $C$ of the tensor $\mathcal{T} \in \mathbb{R}^{I_1 \times I_2 \times \cdots \times I_N}$ provided by TKD with multilinear rank $(R_1, R_2, \cdots, R_N)$ is given by
\begin{equation}
    C = \frac{I_1 I_2 \cdots I_N}{R_1 R_2 \cdots R_N + \sum\limits_{i=1}^{N} I_i R_i}.
\end{equation}
Thus the values of multilinear rank control the compression ratio.

\subsection{Convolutional Layers Compression}
\hspace{10mm} The convolutional operation in CNNs transforms the order-3 input tensor $\mathcal{X} \in \mathbb{R}^{H \times W \times S}$ of feature maps into the order-3 tensor $\mathcal{Y} \in \mathbb{R}^{\hat{H} \times \hat{W} \times \hat{S}}$. The linear mapping is performed with the help of order-4 kernel tensor $\mathcal{W} \in \mathbb{R}^{D \times D \times S \times \hat{S}}$:
\begin{equation}
    \mathcal{Y}_{\hat{h} \hat{w} \hat{s}} = \sum\limits_{i=1}^{D} \sum\limits_{j=1}^{D} \sum\limits_{s=1}^{S} \mathcal{W}_{i j s \hat{s}} \mathcal{X}_{h_i w_j s},
\end{equation}
where $h_i = (\hat{h} - 1)\delta + i - p$ and $w_j = (\hat{w} - 1)\delta + j - p$, $\delta$ is the stride size, $p$ is the zero-padding size.

The spatial dimensions, $D \times D$, of the kernels are typically small, e.g. $3 \times 3$ or $5 \times 5$, in CNNs while filter depth $S$ and number of filters $\hat{S}$ can be big enough. Thus the Tucker decomposition does not have to be performed for all modes of convolutional tensor but only for mode-3 and mode-4. This leads to partial Tucker decomposition of rank-$(R_3, R_4)$ defined:
\begin{equation}
    \mathcal{\widetilde{W}}_{i j s \hat{s}} = \sum\limits_{r_3=1}^{R_3} \sum\limits_{r_4=1}^{R_4} \mathcal{G}_{i j r_3 r_4} \mathbf{U}_{s r_3}^{(3)} \mathbf{U}_{\hat{s} r_4}^{(4)},
\end{equation}
where $\mathcal{G}$ is the core tensor of size $D \times D \times R_3 \times R_4$. The convolution of input $\mathcal{X}$ with kernel in the low-rank Tucker format can be arranged into three consecutive convolutions:
\begin{equation}
    \mathcal{Y}^{(1)}_{h w r_3} = \sum\limits_{s=1}^{S} \mathbf{U}_{s r_3}^{(3)} \mathcal{X}_{h w s} \text{ - pointwise convolution},
\end{equation}
\begin{equation}
    \mathcal{Y}^{(2)}_{\hat{h} \hat{w} r_4} = \sum\limits_{i=1}^{D} \sum\limits_{j=1}^{D} \sum\limits_{r_3=1}^{R_3}  \mathcal{G}_{i j r_3 r_4} \mathcal{Y}^{(1)}_{h_i w_i r_3} \text{ - convolution with the core},
\end{equation}
\begin{equation}
    \mathcal{Y}_{\hat{h} \hat{w} \hat{s}} = \sum\limits_{r_4=1}^{R_4} \mathbf{U}_{\hat{s} r_4}^{(4)} \mathcal{Y}^{(2)}_{\hat{h} \hat{w} r_4} \text{ - pointwise convolution}.
\end{equation}

The described partial TKD for convolutional weights tensors with sufficiently small rank $(R_3, R_4)$ implies compression in terms of reducing number of parameters and reducing number of multiply–accumulate operations (MACs). As the number of parameters in the full convolutional tensor is $D^2 S \hat{S}$, the parameter compression ratio $P$ is expressed in the following way:
\begin{equation}
    P = \frac{D^2 S \hat{S}}{D^2 R_3 R_4 + S R_3 + \hat{S}R_4}.
\end{equation}
The MACs compression ratio, i.e. acceleration, is given by:
\begin{equation}
    M = \frac{D^2 S \hat{S}}{D^2 R_3 R_4 + S R_3 \frac{H W}{\hat{H} \hat{W}} + \hat{S}R_4}.
\end{equation}

\section{Quantization}\label{quant}
\hspace{10mm} The computers can handle just finite numbers of bits to represent the real numbers thus the absolute precision of real numbers representation is unavailable. The conventional approach to manage real values and operations on them in CNNs is to use 32-bit (full precision) or 16-bit (half precision) floating-point arithmetic. However the 32-bit floating point representation often exceeds the required precision. In such cases the smaller number of bits can be used to represent the values. The fixed-point arithmetic with low bitness can be significantly faster with the special hardware support. Another benefit of lower bitness is more efficient cache and registers usage. The quantization also serves for compression purposes. The reduction in memory is straightaway, e.g. changing from 32-bits to 8-bits results in 4x compression in memory.

\hspace{10mm} Quantization of CNNs is divided into two parts – quantization of tensors of weights and quantization of feature maps (activations). In terms of quality restoration after applying quantization the two approaches exists: the post-training quantization which assumes no fine-tuning and quantization-aware training which implies techniques for managing the specificity introduced by quantization during the training stage.

\subsection{Weights}\label{quant_w}
\hspace{10mm} The scheme for quantization of convolutional tensor $\mathcal{W}$ of size $D \times D \times S \times \hat{S}$ consists of the following steps. First, the threshold value $T$ is found. The most straightforward approach is to use the symmetric scheme  \cite{max_quant}, i.e. find the absolute maximum value:
\begin{equation}
    T = \max_{ijs\hat{s}} |\mathcal{W}_{ijs\hat{s}}|.
\end{equation}
The quantization error can be improved if the threshold is found for every filter separately. In this case the threshold $T$ becomes a vector of size $\hat{s}$ and its entries are defined:
\begin{equation}
    T_{\hat{s}} = \max_{ijs} |\mathcal{W}_{ijs\hat{s}}|.
\end{equation}

The distribution of values in the most cases is in the form of a bell with long but sparse tails and symmetric relative to zero value. Thus the absolute maximum as the threshold value is not always the best solution. In such cases the value corresponding to some quantile (e.g. 0.99) can be considered as the threshold. The best quantile can be found via cross-validation procedure. The quantization step is then estimated:
\begin{equation}
    s = \frac{T}{2^{N-1} - 1},
\end{equation}
where $N$ is the number of bits for fixed-point representation.
Finally the quantization of $\mathcal{W}$ is given by:
\begin{equation}\label{eq_w_q}
    {\mathcal{W}}^Q = \text{Clamp}\Big( \Big\lfloor \frac{\mathcal{W}}{s}\Big\rceil, -2^{N-1}, 2^{N-1}-1\Big)s,
\end{equation}
where function Clamp$(x, a, b)$ is defined as:
\begin{equation}
    \text{Clamp}(x, a, b) = 
    \begin{cases}
        a, &\text{if } x \leq a\\
        x, &\text{if } a \leq x \leq b\\
        b, &\text{if } x \geq b.\\
    \end{cases}
\end{equation}

At the fine-tuning stage at the forward pass the weights are quantized according to (~\ref{eq_w_q}). The backward pass uses the method of straight-through estimator \cite{courbariaux2015binaryconnect} and is described in Section ~\ref{quant_a}.

\subsection{Activations}\label{quant_a}
\begin{figure}[t]
    \centering
    \includegraphics[width=0.8\columnwidth]{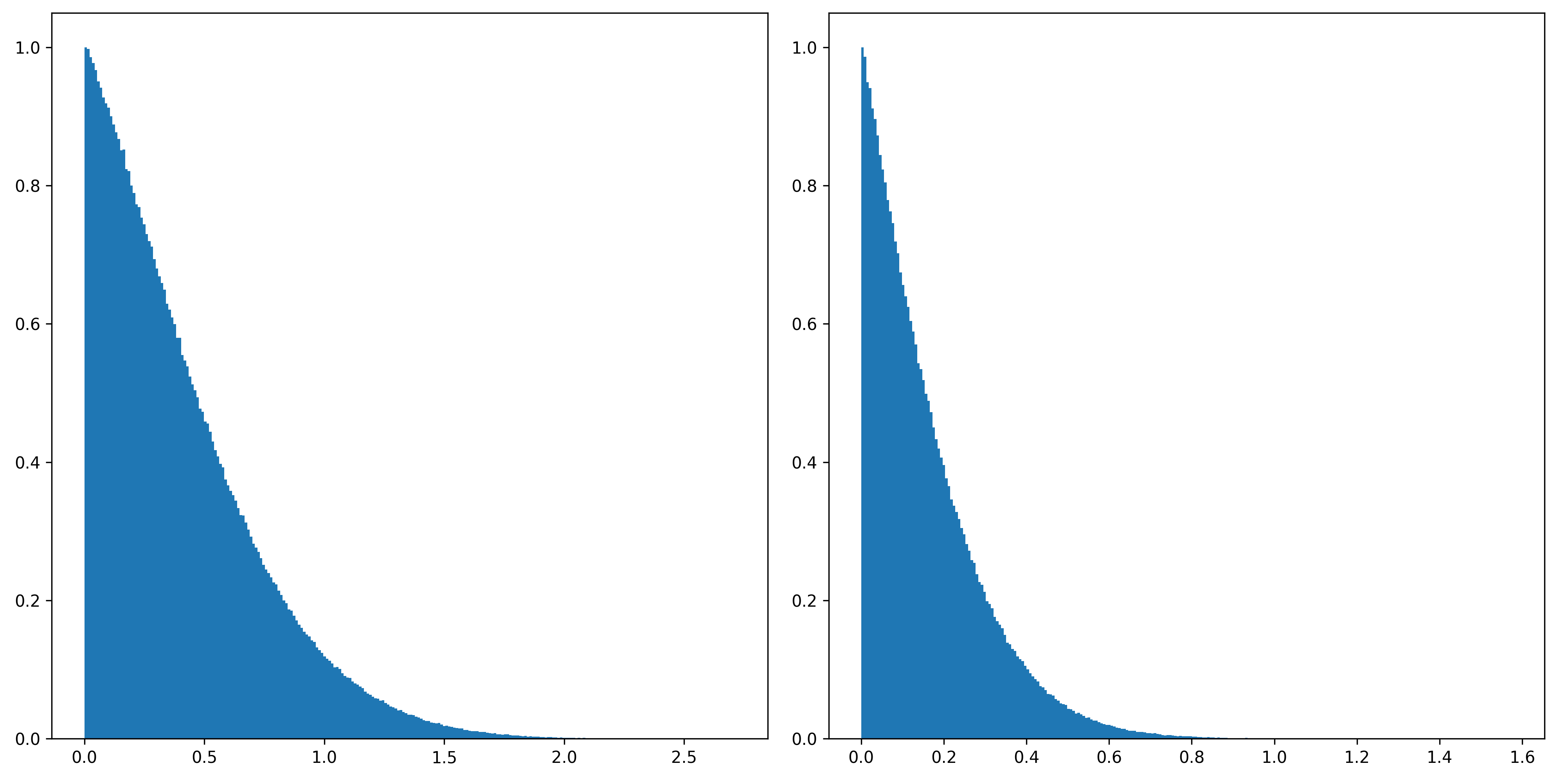}
    \caption{Normalized distribution of activations (excluding zero) of last two layers of ResNet34 for CIFAR-100 classification}
    \label{fig_relu_acts}
\end{figure}

\hspace{10mm} The distribution of feature maps $\mathcal{X}$ strongly depends on the activation function applied. For example, when rectified linear unit (ReLU) activation function is used the distribution has a semi-bell form with peak at zero. In the Figure ~\ref{fig_relu_acts} the normalized distributions of activations (excluding zero point) of the two convolutional layers in the second basic block of ResNet34 for CIFAR-100 classification are shown. In these histograms the activations values belong to $x$-axis and are split into 256 bins while the $y$-axis shows the normalized amount of the values which occur in the bins on the $x$-axis.
Thus the quantization scheme for activations differs from the one for weights. The thresholds for left and right borders are found independently:
\begin{equation}
    T^r = \max_{hws} \mathcal{X}_{hws},
\end{equation}
\begin{equation}
    T^l = \min_{hws} \mathcal{X}_{hws}.
\end{equation}
Then to use the similar notation as for weight quantization the value $T$ is given by:
\begin{equation}
    T = T^r - T^l.
\end{equation}
The quantization step is defined according to the activation function used in the following way:
\begin{equation}
    s = 
    \begin{cases}
        \frac{T}{2^{N} - 1}, &\text{if } \mathcal{X} \succeq 0\\
        \frac{T}{2^{N-1} - 1} &\text{else}.\\
    \end{cases}
\end{equation}
The quantization is also influenced by the activation function:
\begin{equation}
    \mathcal{X}^Q = 
    \begin{cases}
        \text{Clamp}\Big( \Big\lfloor \frac{\mathcal{X}}{s}\Big\rceil, 0, 2^{N}-1\Big)s, &\text{if } \mathcal{X} \succeq 0\\
        \text{Clamp}\Big( \Big\lfloor \frac{\mathcal{X}}{s}\Big\rceil, -2^{N-1}, 2^{N-1}-1\Big)s &\text{else}.\\
    \end{cases}
\end{equation}
The backward pass is defined as:
\begin{equation}
    \Delta = \frac{\partial L}{\partial \mathcal{W}^Q} \odot \mathbf{I}_{x \in S},
\end{equation}
where $L$ is the loss function, $\mathbf{I}_{x \in S}$ is the indicator function, $S = \{x: T^l \leq x \leq T^r\}$, $\odot$ is the element-wise product.

\section{Quality Restoration}\label{qual_rest}
\hspace{10mm} After applying low-rank approximation and quantization the quality of CNN drops. Thus an efficient method for fine-tuning is proposed which is inspired by teacher-student approach.

\hspace{10mm} The initial well-trained cumbersome model in the floating-point representation is assumed as a teacher. The shallow model with convolutional tensors represented in the low-rank Tucker format with fixed-point weights is assumed as a student. The idea of knowledge distillation method introduced in \cite{hinton2015distilling} is applied here for the classification task:
\begin{equation}
    L_{KD} = \alpha \cdot \tau^2 \cdot H(Q^{\tau}_{T}, Q^{\tau}_{S}) + (1-\alpha) \cdot H(Q_{S}, Y_{true}),
\end{equation}
where $H(p, q)$ is the cross-entropy function, $\alpha$ is the hyperparameter, $\tau$ is the temperature, $Q^{\tau}_{T}$ and $Q^{\tau}_{S}$ are teacher's and student's soft targets correspondingly.

The optimization problem is formulated in the following way:
\begin{equation}\label{eq_opt}
\begin{aligned}
\min \quad & \sum\limits_{n=1}^{N} ||\mathcal{W}^{(n)} - \widetilde{\mathcal{W}}^{(n)} ||_F^2 + \lambda||\sigma(\mathcal{Y}^{(n)}) - \sigma(\widetilde{\mathcal{W}}^{(n)}* \widetilde{\mathcal{X}}^{(n)})||_F^2 +  L_{KD}(Q_T, Q_S, \mathcal{Y}_{true})\\
\textrm{s.t.} \quad & \widetilde{\mathcal{W}^{(n)}} \text{is in the low-rank Tucker format;} \\
& \text{factors  of } \widetilde{\mathcal{W}^{(n)}} \text{ are quantized into 1,...,8 bits;} \\
& \text{activations } \widetilde{\mathcal{X}^{(n)}} \text{ are quantized into 1,...,8 bits,}
\end{aligned}
\end{equation}
where $\sigma(\cdot)$ is the non-linear activation function, $\lambda$ is the hyperparameter.
Stochastic gradient descent algorithms are used to optimize the loss function stated in (~\ref{eq_opt}).

\section{Rank Selection}\label{rank_sel}
\hspace{10mm} The appropriate multilinear rank selection is a challenging task. The approach proposed in \cite{kim2015compression} is to use variational Bayesian matrix factorization (VBMF) \cite{Nakajima2013GlobalAS}. Despite that the global analytic VBMF is the prospective method it finds only one multilinear rank for a convolutional tensor of weights, not allowing to control the tradeoff between quality drop and compression ratio. In order to develop the flexible approach for rank selection the following optimization task \cite{tt_phan} for the quantized $n$-th convolutional layer in low-rank format is given by:
\begin{equation}
\begin{aligned}
    \min \quad & R_3^{(n)} + R_4^{(n)} \\
    \textrm{s.t.} \quad & ||\mathcal{Y}^{(n)} - \widetilde{\mathcal{W}}^{(n)}* \widetilde{\mathcal{X}^{(n)}} ||^2_F \leq \epsilon,\\
    & \text{factors  of } \widetilde{\mathcal{W}^{(n)}} \text{ are quantized into 1,...,8 bits,}
\end{aligned}
\end{equation}
where $\widetilde{\mathcal{X}^{(n)}}$ is the quantized input feature map of $n$-th layer, $\mathcal{Y}^{(n)}$ is the output feature maps of $n$-th layer computed with full tensor of weights, i.e., $\mathcal{Y}^{(n)} = \mathcal{W}^{(n)}*\mathcal{X}^{(n)}$.
Rank selection optimization task for compressing the whole CNN with constraints on quality drop is stated similarly to one proposed in \cite{tt_phan} and is given by:
\begin{equation}
\begin{aligned}
    \min \quad & \sum\limits_{n=1}^{N}R_3^{(n)}S^{(n)} + R_4^{(n)} \hat{S}^{(n)} + R_3^{(n)} R_4^{(n)} (D^{(n)})^2 \\
    \textrm{s.t.} \quad & F(\mathcal{Y}) - F(\widetilde{\mathcal{Y}}) \leq t,
\end{aligned}
\end{equation}
where $\mathcal{Y}$ and $\widetilde{\mathcal{Y}}$ are the outputs of the full CNN and the corresponding one with compressed convolutional tensors, $F(\cdot)$ is the objective function which is optimized during training the given CNN.

Opposed the one-shot whole network compression scheme proposed in \cite{kim2015compression} we propose the multistep approach for better compression and convergence at the fine-tuning stage. The single-pass algorithm is summarized in the Algorithm 1.
Another approach for the better fine-tuning is to make multi-pass rank selection, which is described in the Algorithm 2. It outperforms Algorithm 1 as it allows to perform gradual compression of the whole CNN.
\begin{algorithm}[H]\label{alg_1}
\caption{Greedy single-pass rank selection algorithm}
\begin{algorithmic} 
\INPUT Pretrained CNN, set of weights indexes $N$ to compress, set of initial values of ranks $(R_3^n, R_4^n) |_{n \in N}$, set of step values for ranks $(\delta_3^n, \delta_4^n) |_{n \in N}$, metrics function $F()$, metrics threshold $T$, maximum fine-tuning epochs $M$
\For{n in N}
    \State i := 0
    \While{$F(Y) > T$ and i $\leq$ M}
     \State Compute $\widetilde{\mathcal{W}}: \mathcal{W}$ in the rank-$(R_3^n, R_4^n)$ Tucker format with HOOI
     \State Apply quantization on $\widetilde{\mathcal{W}}$
     \State Fine-tune
     \State $R_3^n := R_3^n - \delta_3^n$
     \State $R_4^n := R_4^n - \delta_4^n$
     \State i := i + 1
    \EndWhile
\EndFor
\OUTPUT{Compressed CNN}
\end{algorithmic}
\end{algorithm}

\begin{algorithm}[t!]\label{alg_2}
\caption{Greedy multi-pass rank selection algorithm}
\begin{algorithmic}
\INPUT Pretrained CNN, number of iterations $K$, set of weights indexes $N$ to compress, set of initial values of ranks $(R_3^n, R_4^n) |_{n \in N}$, set of step values for ranks $(\delta_3^n, \delta_4^n) |_{n \in N}$, metrics function $F()$, metrics threshold $T$, maximum fine-tuning epochs $M$
\For{k=1:K}
    \For{n in N}       
         \State Compute $\widetilde{\mathcal{W}}: \mathcal{W}$ in the rank-$(R_3^n, R_4^n)$ Tucker format with HOOI
         \State Apply quantization on $\widetilde{\mathcal{W}}$
         \State Fine-tune
         \If {F(Y) < T}
            \State \textbf{break}
         \EndIf
         \State $R_3^n := R_3^n - \delta_3^n$
         \State $R_4^n := R_4^n - \delta_4^n$
    \EndFor
\EndFor
\OUTPUT{Compressed CNN} 
\end{algorithmic}
\end{algorithm}

\section{Results}\label{results}
\hspace{10mm} In order to verify the efficiency of the proposed compression and acceleration approach we conducted experiments with ResNet18 and ResNet34 CNNs on CIFAR-10, CIFAR-100 and Imagenet datasets for classification tasks. The CIFAR-10 dataset consists of 60000 32x32 colour images in 10 classes, with 6000 images per class. The classes are all mutually exclusive. The CIFAR-100 contains the same images as the CIFAR-10, except it has 100 classes and so 600 images per class. The Imagenet-1k (ILSVRC) contains approximately 1 million images of the real world and 1000 object classes. As the 8-bit fixed-point representation is supported in many inference frameworks and can be efficiently implemented in almost any hardware the detailed experiments were conducted with 8-bit quantization.

\subsection{CIFAR-10 Experiments}\label{res_c10}
\begin{figure}[!h]
  \begin{subfigure}{\linewidth}
  \includegraphics[width=.5\linewidth]{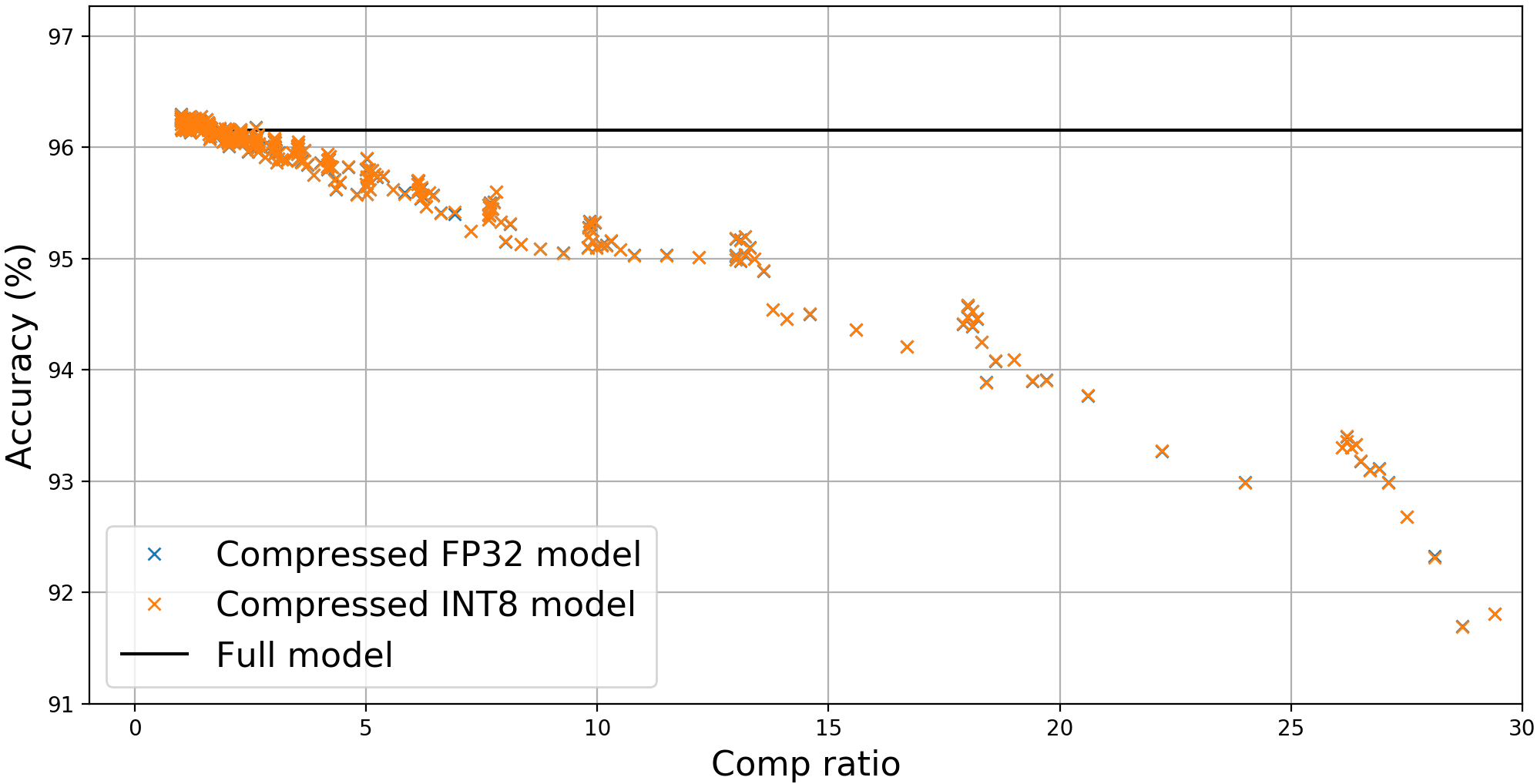}
  \hfill
  \includegraphics[width=.5\linewidth]{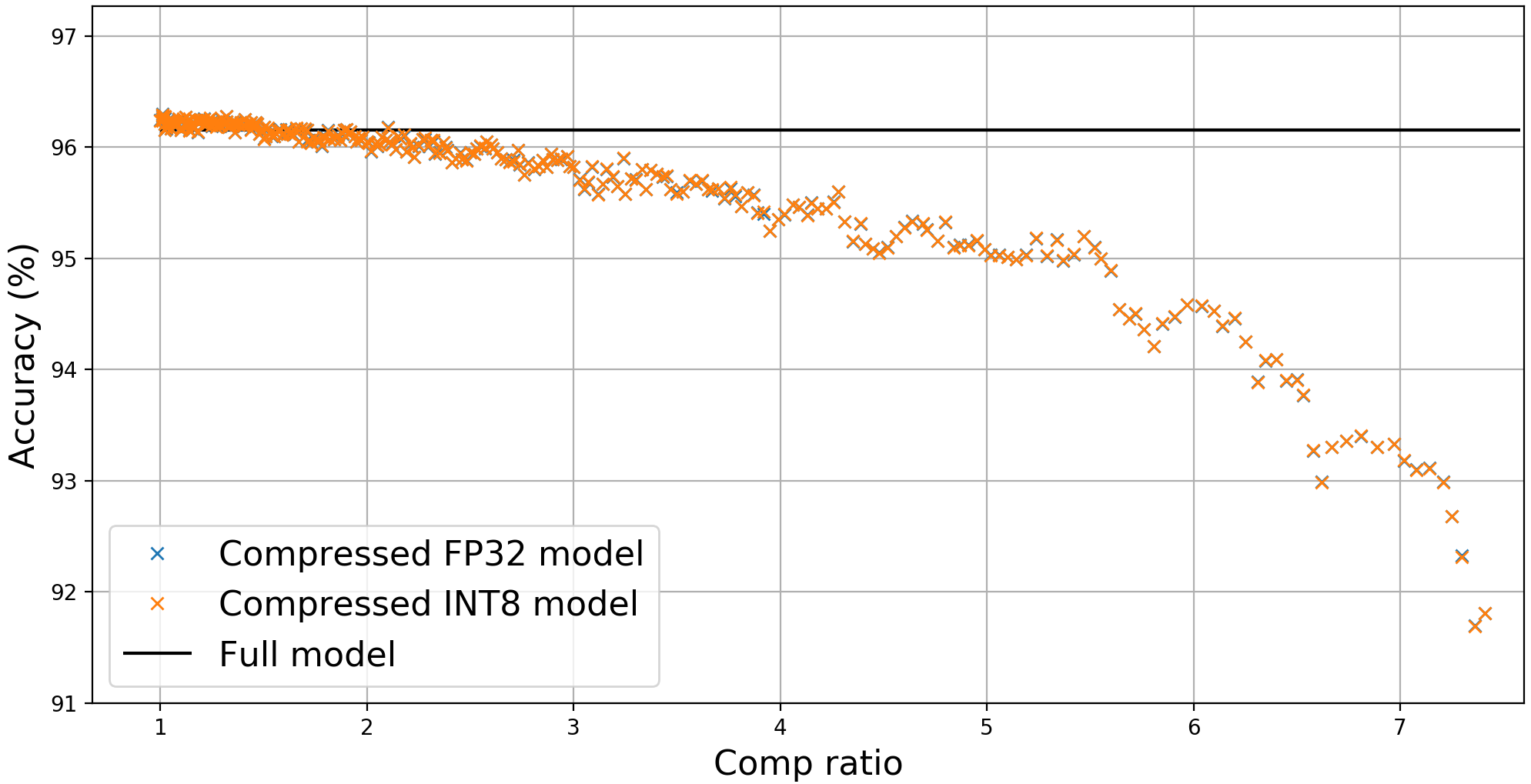}
  \end{subfigure}
\caption{ResNet18 on CIFAR-10. Left: accuracy vs parameters compression ratio. Right: accuracy vs GMACs compression ratio}
\label{r18_c10_head}
\end{figure}

\hspace{10mm} In the Figure ~\ref{r18_c10_head} one can see the dependency of accuracy drop according to the compression ratio for ResNet18 on CIFAR-10 datset. Each dot in the plots depicts the test accuracy of the compressed model with unique configuration of multilinear ranks of compressed layers. The noticeable fact that at the small compression ratios the accuracy of the compressed model is even slightly higher than the one of the full model.

\begin{figure}[!h]
  \begin{subfigure}{\linewidth}
  \includegraphics[width=.5\linewidth]{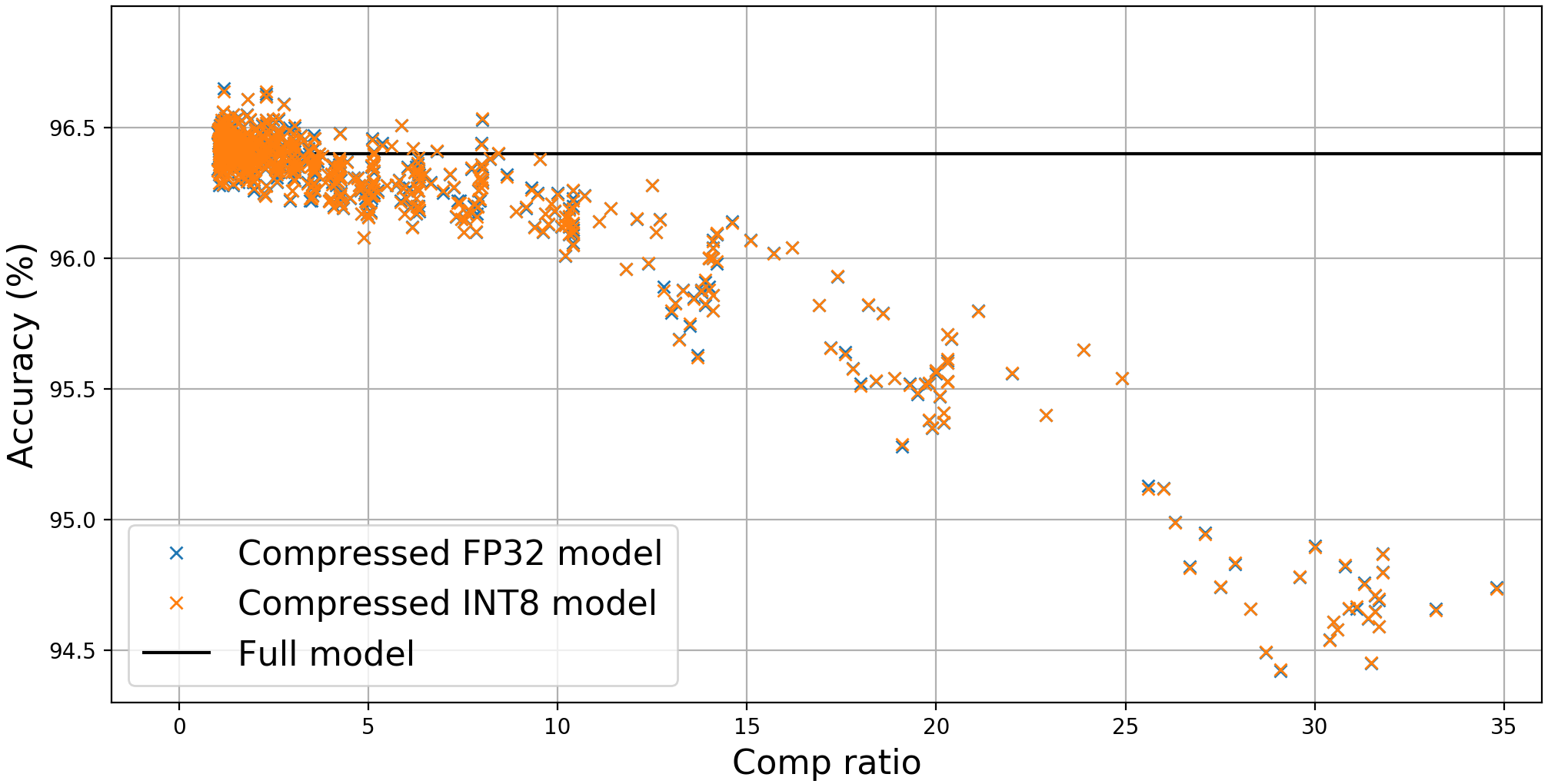}
  \hfill
  \includegraphics[width=.5\linewidth]{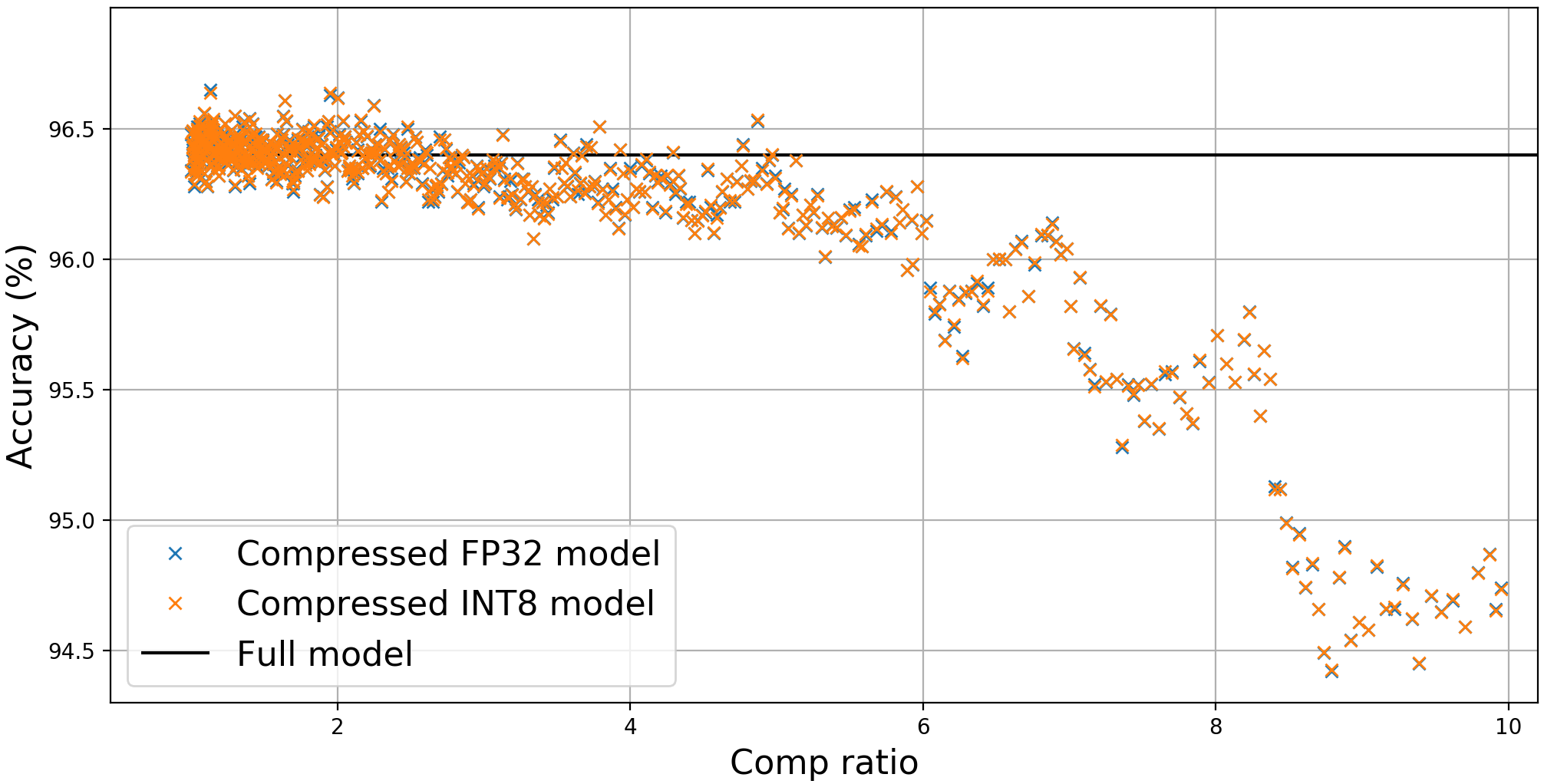}
  \end{subfigure}
\caption{ResNet34 on CIFAR-10. Left: accuracy vs parameters compression ratio. Right: accuracy vs GMACs compression ratio}
\label{r34_c10_head}
\end{figure}

\hspace{10mm} Quite similar plots are received for ResNet34 on CIFAR-10 which are depicted in the Figure ~\ref{r34_c10_head}.
Both ResNet18 and ResNet34 showed no accuracy drop comparing floating-point and 8-bit fixed-point representations for each compressed model. Thus the real memory compression is 4 times bigger if 8-bit quantization is applied.

\newpage
\subsection{CIFAR-100 Experiments}\label{res_c100}
\begin{figure}[!h]
  \begin{subfigure}{\linewidth}
  \includegraphics[width=.5\linewidth]{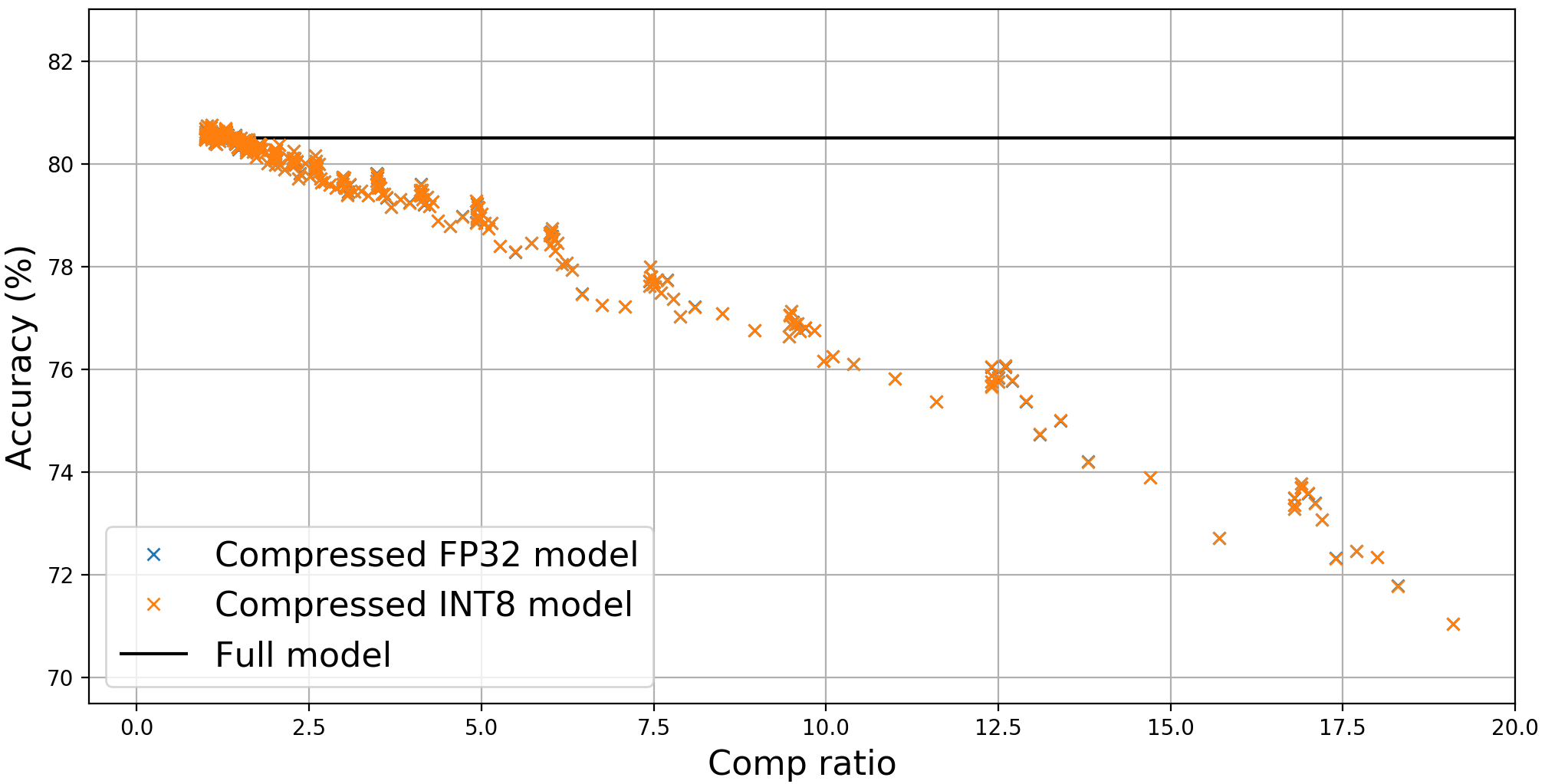}
  \hfill
  \includegraphics[width=.5\linewidth]{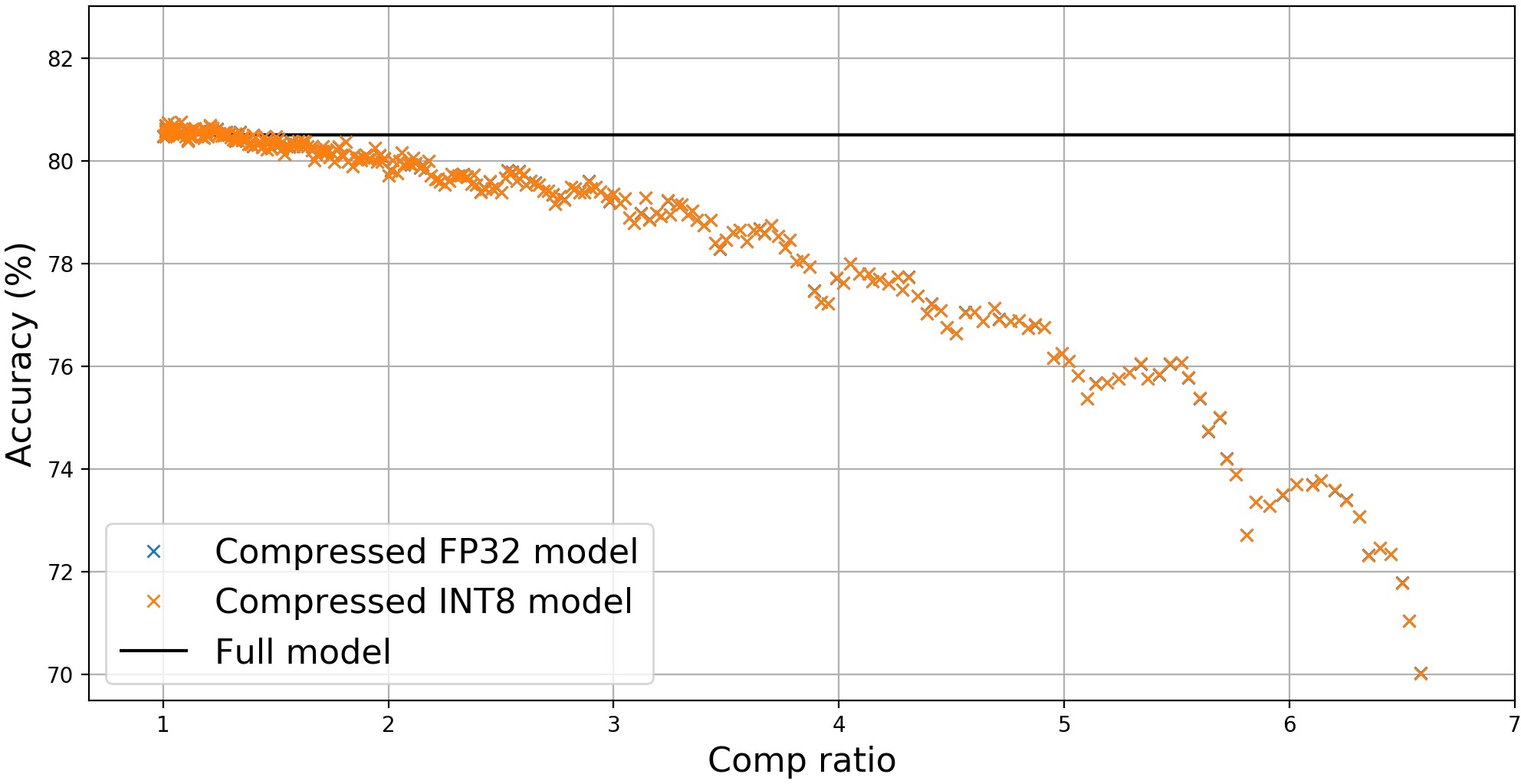}
  \end{subfigure}
\caption{ResNet18 on CIFAR-100. Left: accuracy vs parameters compression ratio. Right: accuracy vs GMACs compression ratio}
\label{r18_c100_head}
\end{figure}

\hspace{10mm} The Figure ~\ref{r18_c100_head} shows the results for ResNet 18 CIFAR-100. The CIFAR-100 is much more complicated task than CIFAR-10 so the efficient compression ratios are smaller than for CIFAR-10. Still the compressed models produce slightly higher test accuracy compared to the full-precision model.

\begin{figure}[!h]
  \begin{subfigure}{\linewidth}
  \includegraphics[width=.5\linewidth]{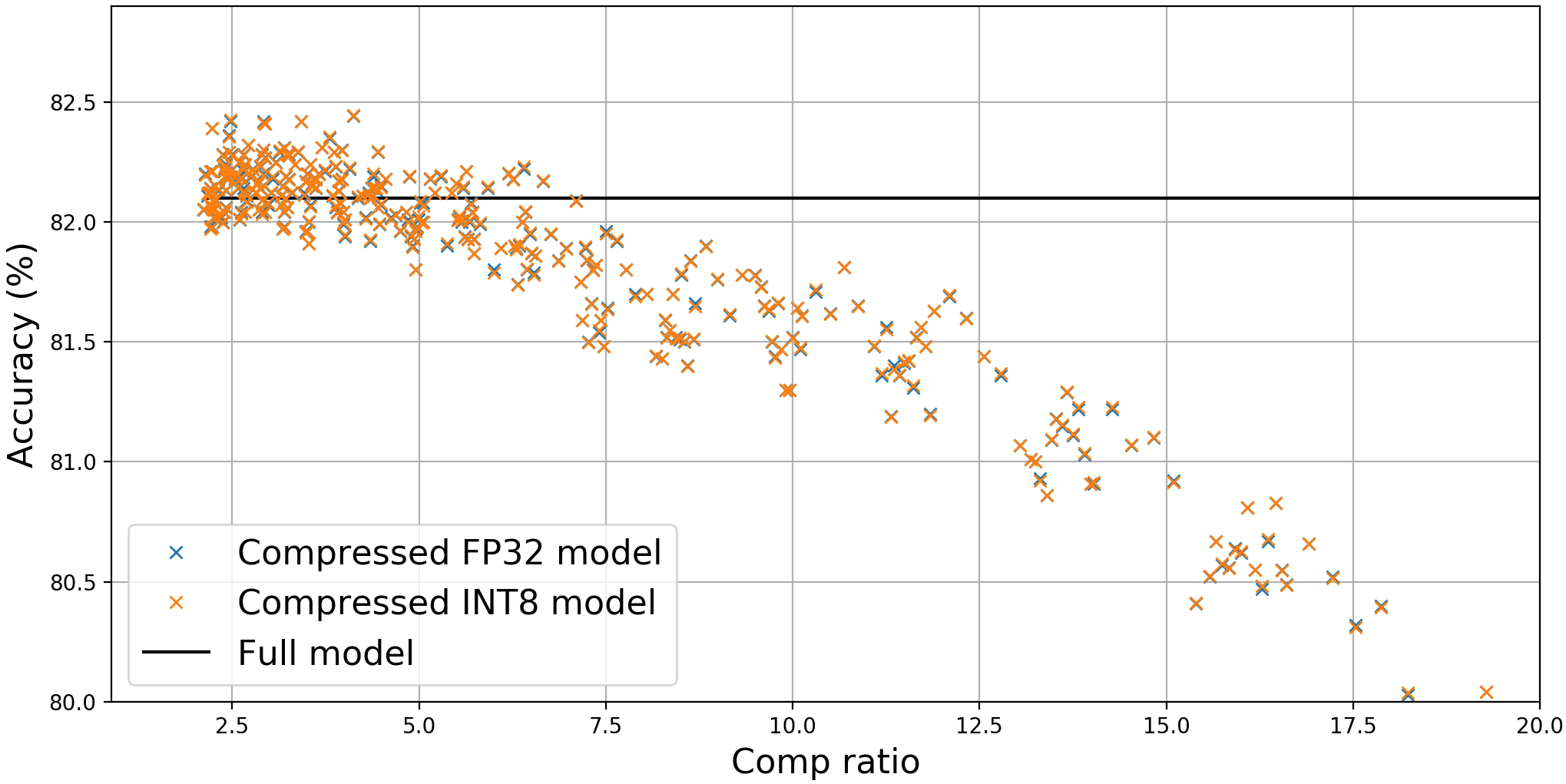}
  \hfill
  \includegraphics[width=.5\linewidth]{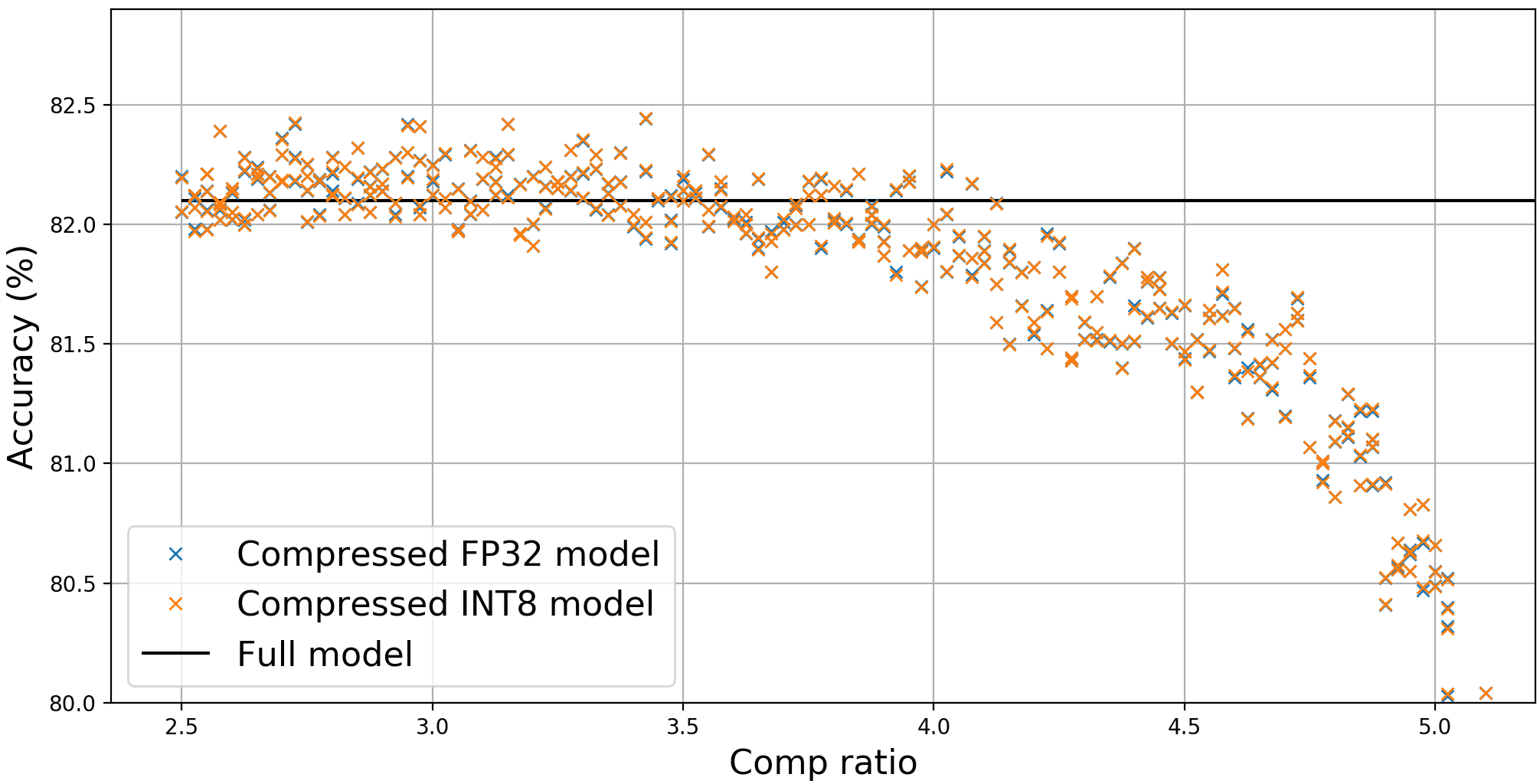}
  \end{subfigure}
\caption{ResNet34 on CIFAR-100. Left: accuracy vs parameters compression ratio. Right: accuracy vs GMACs compression ratio}
\label{r34_c100_head}
\end{figure}

The results for ResNet34 on CIFAR-100 are shown in the Figure ~\ref{r34_c100_head}. The pictures are rather similar with the ones for CIFAR-10. The 8-bit quantization shows its perfect results with no relative accuracy drop for both ResNet18 and ResNet34 on CIFAR-100.

\newpage
\subsection{Imagenet Experiments}\label{res_imnet}
\begin{figure}[!h]
  \begin{subfigure}{\linewidth}
  \includegraphics[width=.5\linewidth]{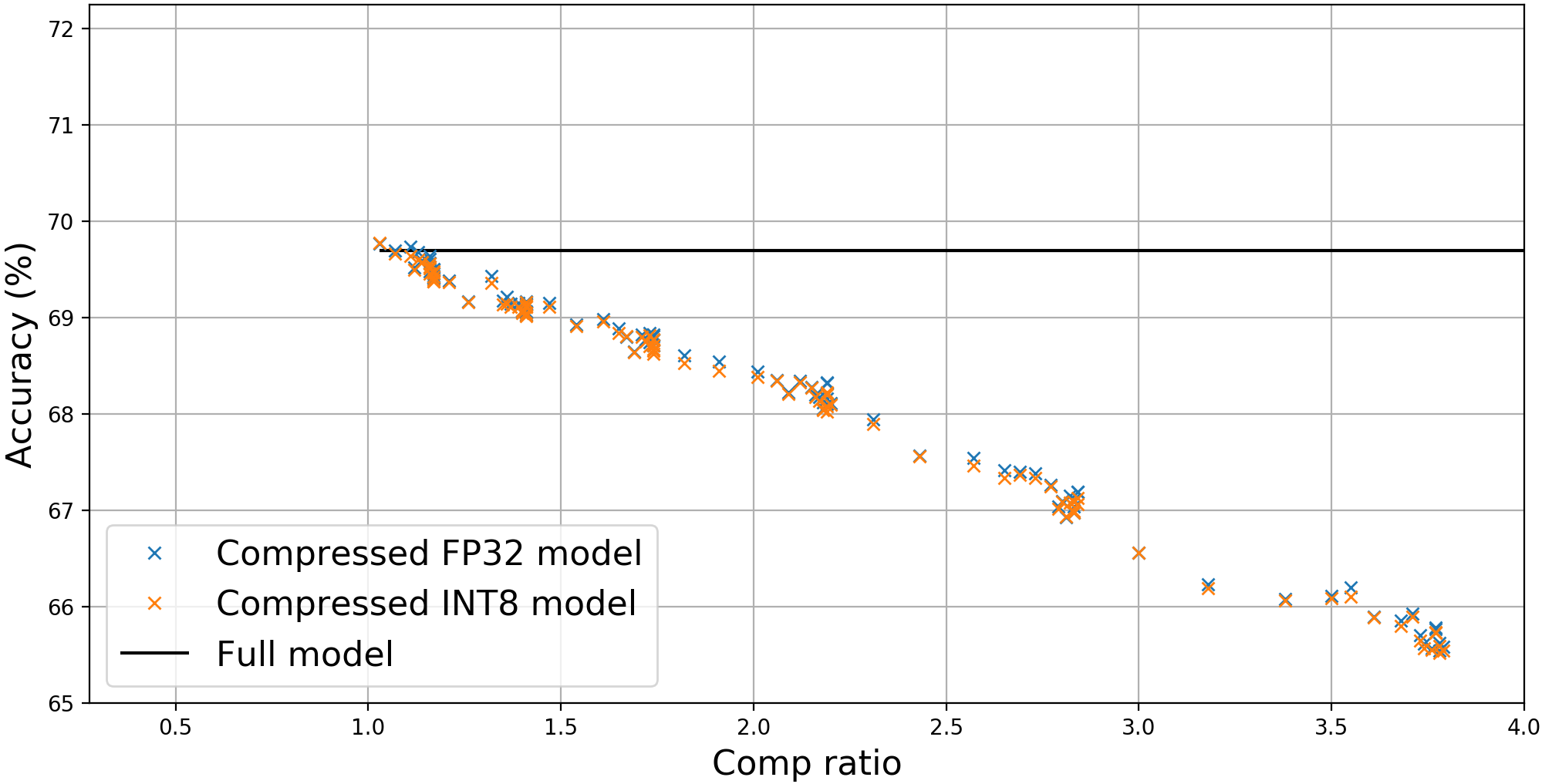}
  \hfill
  \includegraphics[width=.5\linewidth]{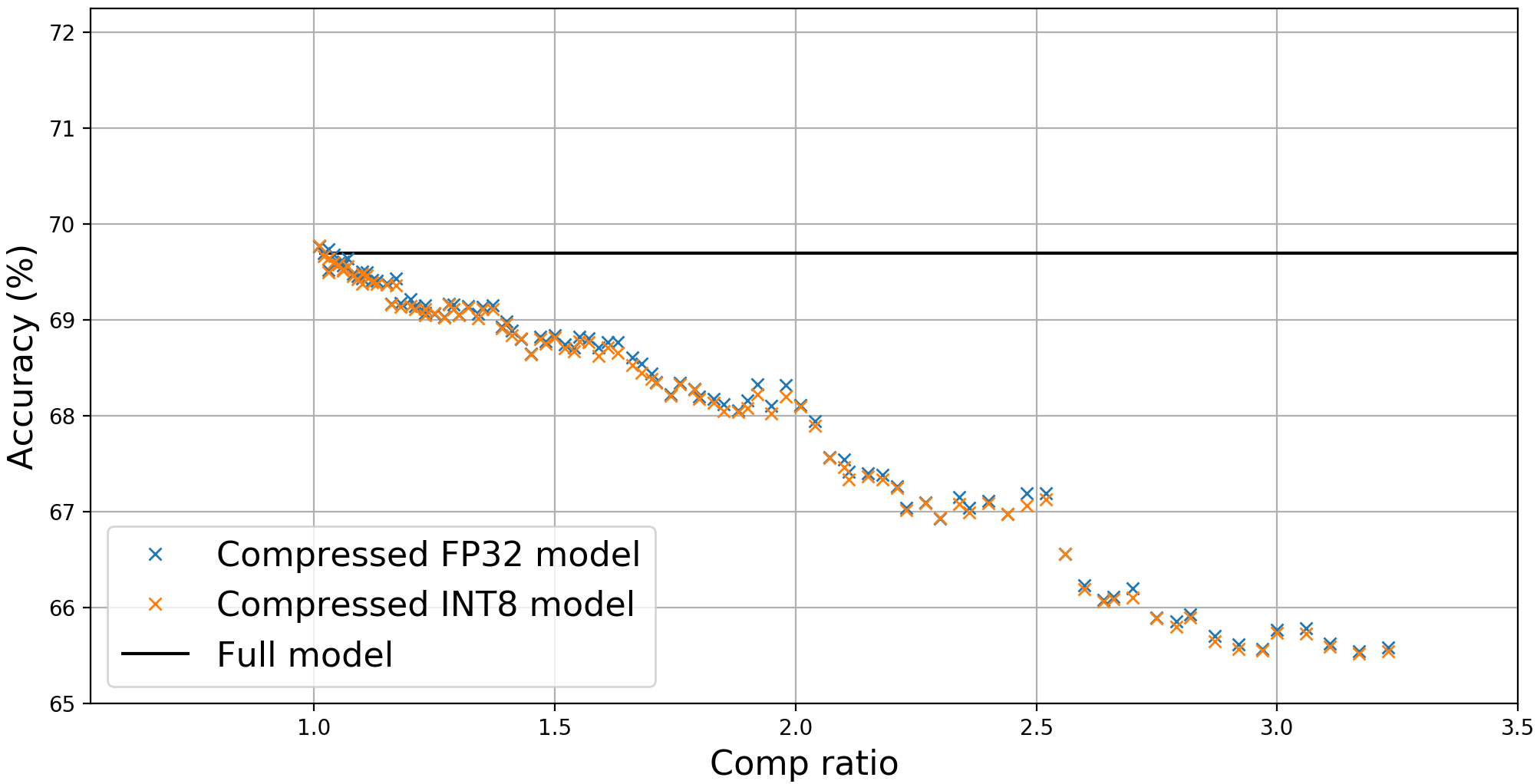}
  \end{subfigure}
\caption{ResNet18 on Imagenet. Left: accuracy vs parameters compression ratio. Right: accuracy vs GMACs compression ratio}
\label{r18_imnet}
\end{figure}

\hspace{10mm} The results of experiments with ResNet18 on Imagenet dataset are depicted in the Figure ~\ref{r18_imnet}. The ILSVRC is very complicated task so the big compression ratios without accuracy drop are unavailable for ResNet18 - the smallest one among ResNets for Imagenet. However the 8-bit quantized models show almost the same test accuracy compared to the corresponding full-precision ones.

\subsection{Low Bitness Study}\label{dif_bits}
\begin{figure}[!h]
    \centering
    \includegraphics[width=0.70\columnwidth]{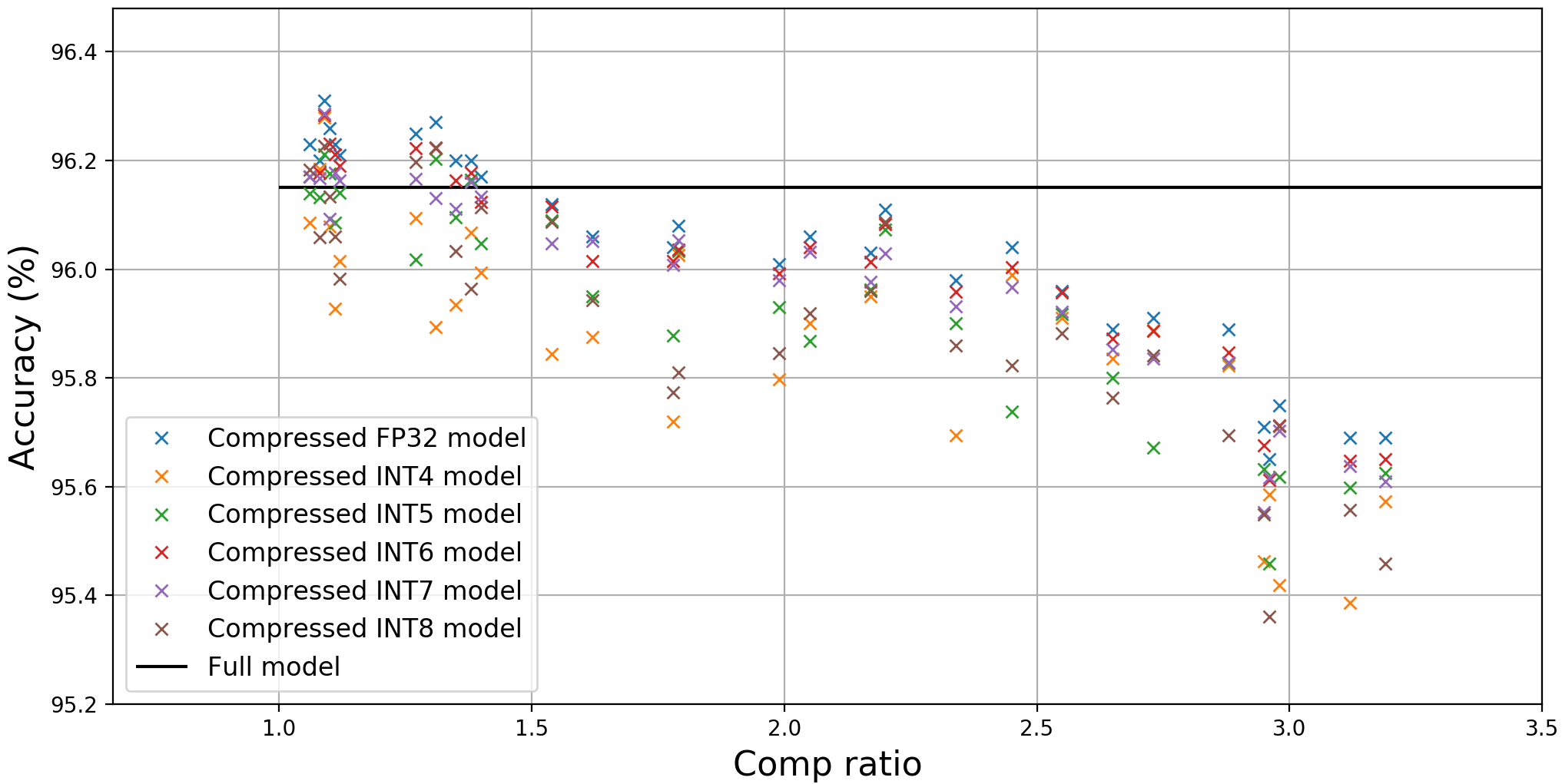}
    \caption{Accuracy vs GMACs compression ratio for 4, 5, 6, 7 and 8 bits quantization. ResNet18 on CIFAR-10}
    \label{r18_c10_bits}
\end{figure}

\hspace{10mm} The experiments with a different number of bits were conducted for ResNet18 on CIFAR-10 dataset and the results are shown in Figure ~\ref{r18_c10_bits}. The appropriate fine-tuning methods allowed to achieve almost no accuracy drop for all quantized models.

\section{Comparison with Other Methods}\label{comparison}
\hspace{10mm} The first method for comparison is based on singular value decomposition (SVD) of weights \cite{denil2013predicting}. The convolutional tensor $\mathcal{W} \in \mathbb{R}^{D \times D \times S \times \hat{S}}$ is firstly reshaped into matrix $\mathbf{W}$ of shape $D^2S \times \hat{S}$. Then the rank-$R$ truncated SVD is applied:
\begin{equation}\label{eq_w_svd}
    \mathbf{W} = \mathbf{U} \mathbf{S} \mathbf{V}^T.
\end{equation}
The initial convolution is split into consecutive two. The fisrt factor $\mathbf{W}^{(1)} = \mathbf{U} \mathbf{S}^{\frac{1}{2}}$ is reshaped into the tensor $\mathcal{W}^{(1)}$ of size $D \times D \times S \times R$ used for the first convolution while the second factor $\mathbf{W}^{(2)} = \mathbf{S}^{\frac{1}{2}} \mathbf{V}^T$ is reshaped into $\mathcal{W}^{(2)}$ of size $1 \times 1 \times R \times \hat{S}$ and corresponds to the pointwise convolution. In result the convolutional tensor in low-rank format is given by:
\begin{equation}
    \widetilde{\mathcal{W}}_{i_h i_w i_s i_{\hat{s}}} = \sum\limits_{i_r=1}^{R} \mathcal{W}_{i_h i_w i_s i_r}^{(1)} \mathbf{W}_{i_r i_{\hat{s}}}^{(2)}.
\end{equation}

\begin{figure}[!t]
    \centering
    \includegraphics[width=0.8\columnwidth]{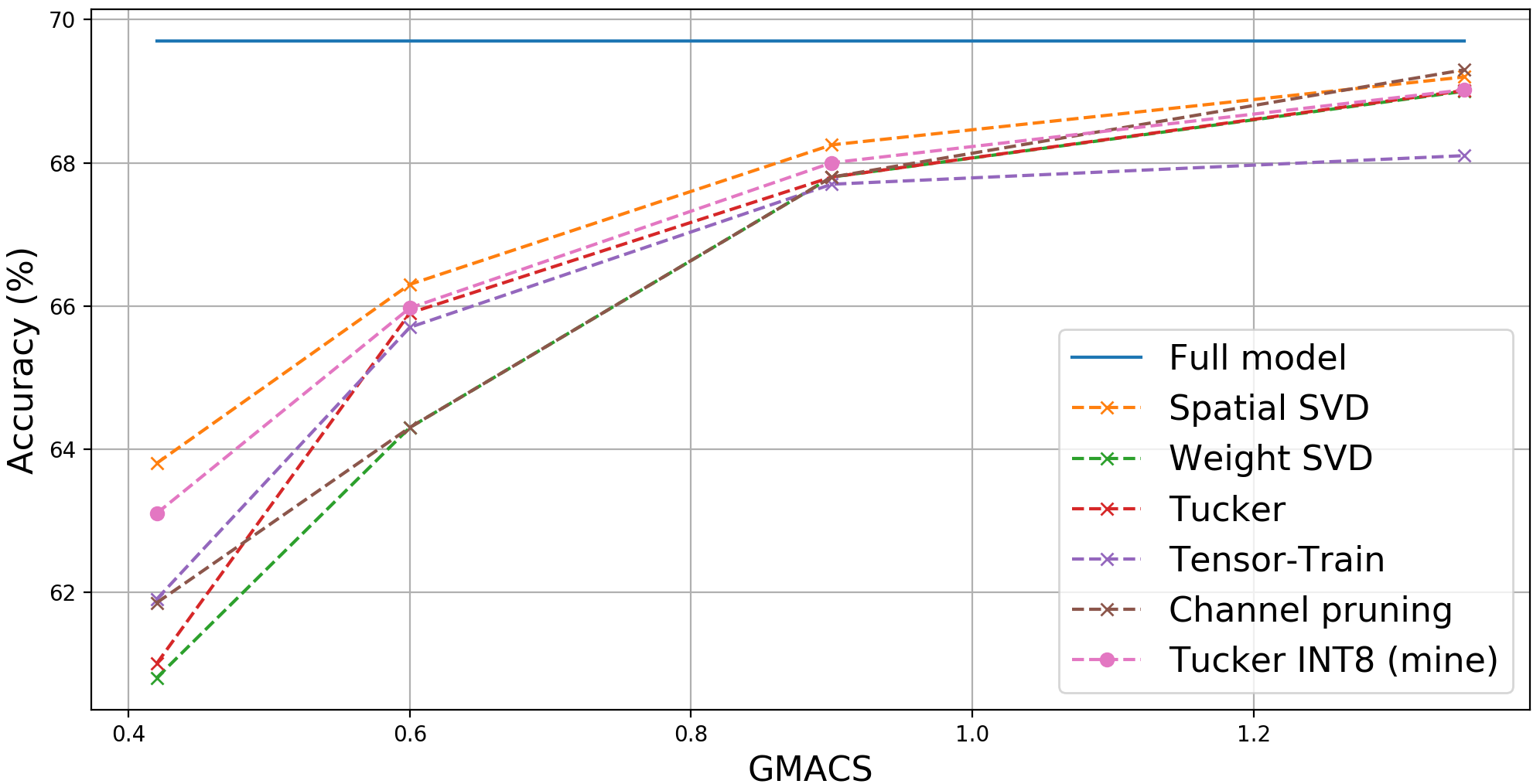}
    \caption{Methods comparison. ResNet18 on Imagenet}
    \label{r18_imnet_comparison}
\end{figure}

\hspace{10mm} Another approach based on low-rank approximation was proposed in \cite{tai2015convolutional}. In this method the weights tensor $\mathcal{W} \in \mathbb{R}^{D \times D \times S \times \hat{S}}$ is reshaped into matrix $\mathbf{W}$ of shape $DS \times D\hat{S}$.
The result of rank-$R$ truncated SVD is the low-rank tensor given by:
\begin{equation}
    \widetilde{\mathcal{W}}_{i_s i_h i_{\hat{s}} i_w} = \sum\limits_{i_r=1}^{R} \mathcal{W}_{i_s i_h i_r}^{(1)} \mathbf{W}_{i_r i_{\hat{s}} i_w}^{(2)},
\end{equation}
where $\mathcal{W}^{(1)}$ is used for the convolution with vertical filter of size $d \times 1$, $\mathcal{W}^{(2)}$ - for convolution with horizontal filter of size $1 \times d$. This method can be called spatial SVD based as it utilized the redundancy in channels.

\hspace{10mm} We also used the method based on pure Tucker decomposition as was proposed in \cite{kim2015compression}. 

\hspace{10mm} The tensor-train decomposition was used in \cite{garipov2016ultimate} for low-rank representation. Garipov et al. proposed to reorder the convolutional tensor as $\mathcal{W} \in \mathbb{R}^{S \times D \times S \times \hat{S}}$ and apply rank-$(R_1, R_2, R_3)$ tensor-train decomposition:
\begin{equation}
    \widetilde{\mathcal{W}}_{i_s i_h i_w i_{\hat{s}}} = \sum\limits_{i_{r_1}=1}^{R_1} \sum\limits_{i_{r_2}=1}^{R_2} \sum\limits_{i_{r_3}=1}^{R_3} \mathbf{W}_{i_s i_{r_1}}^{(1)} \mathcal{W}_{i_{r_1} i_h i_{r_2}}^{(2)} \mathcal{W}_{i_{r_2} i_w i_{r_2}}^{(3)} \mathbf{W}_{i_{r_3} i_{\hat{s}}}^{(4)}.
\end{equation}

\hspace{10mm} The compression method based on lasso feature selection for channel pruning was proposed in \cite{he2017channel}. Authors formulated the optimization task in the following way:
\begin{equation}
\begin{aligned}
    \argmin_{\beta, \mathbf{W}} \quad & \Big|\Big|\mathbf{Y} - \sum_{i=1}^{s} \beta_i \mathbf{X}_i \mathbf{W}_i^T  \Big|\Big|_F^2 \\
    \textrm{s.t.} \quad & ||\beta||_0 \leq \hat{t},
\end{aligned}
\end{equation}
where $\mathbf{X} \in \mathbb{R}^{n \times s \times d \times d}$ is the input batch sampled from the corresponding feature maps of the uncompressed model, $\mathbf{X}_i \in \mathbb{R}^{n \times d^2}$ is the concatenation of $n$ samples in $i$-th channel, $\mathbf{W}_i \in \mathbb{R}^{\hat{s} \times d^2}$ is the corresponding $i$-th channel of the filter, $\beta$ is the vector of coefficients, $\hat{t} \leq s$ is the desired number of channels. The $L_0$ norm is relaxed so that the optimization is reformulated in the form:
\begin{equation}
\begin{aligned}
    \argmin_{\beta, \mathbf{W}} \quad & \Big|\Big|\mathbf{Y} - \sum_{i=1}^{s} \beta_i \mathbf{X}_i \mathbf{W}_i^T  \Big|\Big|_F^2 + \lambda ||\beta||_1 \\
    \textrm{s.t.} \quad & ||\mathbf{W}_i||_F = 1,
\end{aligned}
\end{equation}
which results in lasso regularization.

\hspace{10mm} The results of comparative experiments with other compression methods of ResNet18 for Imagenet classification are shown in Figure ~\ref{r18_imnet_comparison}. The proposed compression method outperforms all other methods almost everywhere except Spatial SVD based. However our models are additionally quantized into 8 bits which makes compression more efficient comparing to all other methods.

\section{Conclusion}\label{concl}
\hspace{10mm} To summarize the obtained results we indicate the key aspects. The proposed approach of combining low-rank tensor approximation and quantization shows its high potential. The method for fast quality restoration after applying tensor decomposition and quantization is proposed. Without specific fine-tuning techniques the convergence is much slower and quite unstable. Moreover it is shown that the tradeoff between accuracy drop and compression ratio can be controlled. This is important feature as different requirements for different task may exist so that flexible approach is convenient in many cases. Such flexibility can also be very useful when the same well-trained model must be deployed on different devices with different hardware requirements. Another competitive feature of the proposed method is end-to-end fine-tuning which speeds up and simplifies the application of the method for the new tasks and models. Our technique can thus be implemented as a part of a deep learning framework and be offered as out-of-the-box solution for compression and acceleration of the CNNs. However the process of finding the configuration for the best compression or acceleration although almost fully automated can still be time-consuming. This is the main limitation of the proposed approach.

\hspace{10mm} The experiments with deep reinforcement learning (DRL) for efficient rank selection should be developed as a part of the future work. We also plan to develop methods for applying extreme quantization bitness, i.e. binary and ternary, combinig with low-rank tensor approximation.

\bibliography{references}  






\end{document}